\documentclass{article}

\PassOptionsToPackage{numbers, sort&compress}{natbib}
\usepackage[preprint,dandb]{neurips_2025}

\usepackage[utf8]{inputenc} 
\usepackage[T1]{fontenc}    
\usepackage{hyperref}       
\usepackage{url}            
\usepackage{booktabs}       
\usepackage{amsfonts}       
\usepackage{nicefrac}       
\usepackage{microtype}      
\usepackage{xcolor}         
\usepackage{colortbl}      
\usepackage{graphicx}
\usepackage{amsmath}
\usepackage{amssymb}
\usepackage{multirow}
\usepackage{subcaption}
\usepackage{makecell}

\usepackage[export]{adjustbox}

\title{The Catechol Benchmark: Time-series Solvent Selection Data for Few-shot Machine Learning}

\author{%
  Toby Boyne\textsuperscript{1}\thanks{\texttt{t.boyne23@imperial.ac.uk} ; \textsuperscript{\textdagger}\texttt{jose@solvechemistry.com}}, Juan S. Campos\textsuperscript{1}, Becky D. Langdon\textsuperscript{1}, Jixiang Qing\textsuperscript{1}, Yilin Xie\textsuperscript{1}\\ 
  \textbf{Shiqiang Zhang\textsuperscript{1}, Calvin Tsay\textsuperscript{1}, Ruth Misener\textsuperscript{1}, Daniel W. Davies\textsuperscript{2}, Kim E. Jelfs\textsuperscript{2}} \\
  \textbf{Sarah Boyall\textsuperscript{3}, Thomas M. Dixon\textsuperscript{3}, Linden Schrecker\textsuperscript{3}, Jose Pablo Folch\textsuperscript{3}\textsuperscript{\textdagger}} \\
  Department of Computing, Imperial College London, London, UK\textsuperscript{1} \\
  Department of Chemistry, Imperial College London, London, UK\textsuperscript{2} \\
  SOLVE Chemistry, London, UK\textsuperscript{3}
}

\DeclareMathOperator*\argmax{arg\,max}

\begin{document}

\maketitle

\begin{abstract}
  Machine learning has promised to change the landscape of laboratory chemistry, with impressive results in molecular property prediction and reaction retro-synthesis. However, chemical datasets are often inaccessible to the machine learning community as they tend to require cleaning, thorough understanding of the chemistry, or are simply not available. In this paper, we introduce a novel dataset for yield prediction, providing the first-ever transient flow dataset for machine learning benchmarking, covering over 1200 process conditions. While previous datasets focus on discrete parameters, our experimental set-up allow us to sample a large number of continuous process conditions, generating new challenges for machine learning models. We focus on solvent selection, a task that is particularly difficult to model theoretically and therefore ripe for machine learning applications. We showcase benchmarking for regression algorithms, transfer-learning approaches, feature engineering, and active learning, with important applications towards solvent replacement and sustainable manufacturing.
\end{abstract}

\section{Introduction}

Machine learning (ML) and artificial intelligence (AI) have showcased enormous potential in empowering the world of the natural sciences: from famous examples such as AlphaFold  for protein predictions~\citep{jumper2021highly}, to fusion reactor control~\citep{degrave2022magnetic}, disease detection \citep{zhou2023foundation}, battery design \citep{folch2023combining}, and material discovery \citep{raccuglia2016machine}, among many more. However, we seldom see the machine learning community benchmark new methods in physical science datasets, mostly due to the difficulty in cleaning real-world data, the need for interdisciplinary understanding to correctly benchmark, and most importantly, how expensive the data can be to produce, resulting in many datasets being locked behind closed doors by large companies.

AIchemy (\url{https://aichemy.ac.uk}) is an interdisciplinary UK hub with the mission of transforming the chemistry-AI interface via aiding the collaboration of chemists and AI researchers, as well as addressing gaps in data standards, curation, and availability for AI use. In partnership with SOLVE Chemistry (\url{https://www.solvechemistry.com}), we present a first important step into addressing the dataset gap with the introduction of a new and unique open dataset for benchmarking low-data machine learning algorithms for chemistry.

Solvent selection is one of the biggest challenges for chemical manufacturing, with solvents often being the main source of waste in the manufacturing process \citep{constable2007perspective}. Increased regulation on solvents and a drive to making process manufacturing more sustainable led to an interest in the discovery of greener solvents and for improved solvent replacement tools. However, most of the solvent replacement tools focus purely on learning unsupervised representations of solvents, with the hope that experimentalists can find solvents with similar properties to replace those with environmental concerns. A much stronger approach would consider the interaction of a variety of different solvents with a reaction of interest to directly predict reaction yields, in such a way that the best possible solvent can be selected according to a yield-sustainability trade-off.

Machine learning approaches have been shown to be a powerful tool for the prediction of chemical reaction conditions. Success has been reported in retro-synthesis \citep{karpov2019transformer, tetko2020state}, condition recommendations \citep{gao2018using}, product predictions \citep{coley2019graph, tu2022permutation}, among others. While yield prediction has proven to be more difficult due to large inconsistencies in procedure and data reporting \citep{wigh2024orderly}, we have still seen promising yield prediction results for smaller and more carefully curated datasets \citep{schwaller2021prediction, griffiths2023gauche, rankovic2024bayesian, rankovic2025gollum}. However, these datasets lack the continuous reaction conditions, such as temperature and residence time, that are required to scale-up processes to practical manufacturing conditions. 

\begin{figure}
  \centering
  \includegraphics[width = 0.8\textwidth]{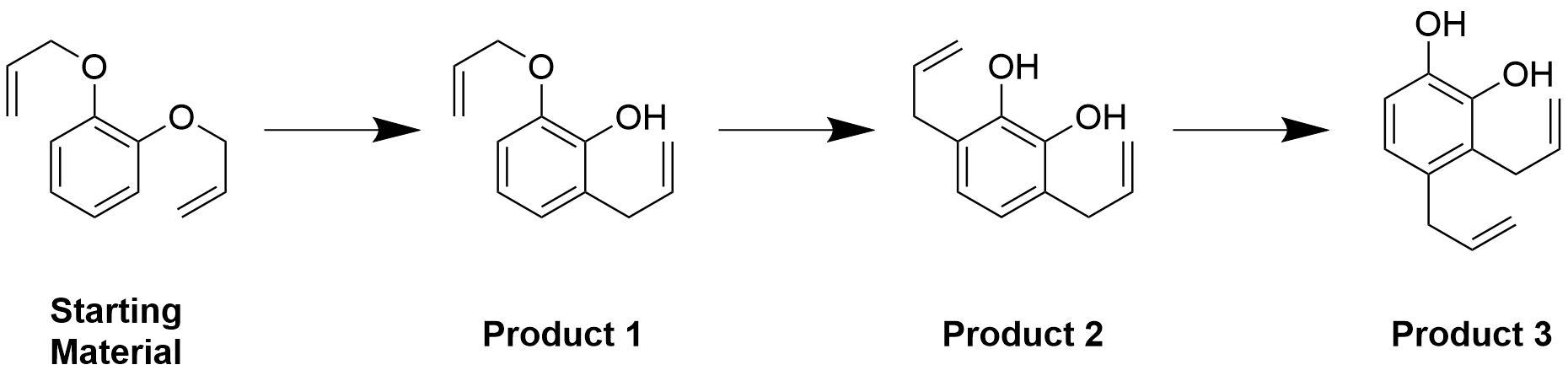}
  \caption{Data was gathered on the rearrangement of allyl substituted catechol. By subjecting the reaction mixture to high temperatures, we begin a cascade reaction forming multiple rearrangement products. We investigate the yield of the reaction for a range of different solvents. Product 1 was not observed and reacted immediately to form Product 2 and later 3.}
  \label{fig: reaction_illustration}
\end{figure}

In this paper, we release the first machine-learning-ready transient flow dataset, a framework that allows for quick and efficient screening of continuous reaction conditions. We specifically provide yield data over the uni-molecular allyl substituted catechol reaction, shown in Figure \ref{fig: reaction_illustration}, with dense measurements across the residence time, temperature, and solvent space. We answer the call for more flow chemistry reaction data \citep{deadman2025wanted}, further showcase how this type of \textit{kinetic data} poses new challenges to current machine learning methods for chemistry, and identify potential solutions.

\subsection{Related works}

Reaction datasets are common in chemistry research, but their suitability for machine learning benchmarking tends to be poor. This can be a result of improper formatting or documentation, incomplete information about reaction conditions or the experimental set-up, or the lack of machine readability, leading to limited usage by the ML community. However, some effort has been made to address this, with the biggest example being the creation of the Open Reaction Database (ORD) \citep{kearnes2021open}, a repository containing over 2M different reactions, many of which come from US patent data (USPTO) \citep{lowe2012extraction}. However, the dataset falls short in some aspects, in particular with respect to machine learning readiness and data inconsistencies across reactions. 

ORDerly \citep{wigh2024orderly} allows for easy cleaning and preparation of ORD data, showing the promise of the dataset for forward and retro-synthetic prediction using transformers; however, it also shows that yield prediction cannot be done well due to data inconsistencies. \citet{schwaller2021prediction} drew similar conclusions when using the USPTO dataset, stating that reaction conditions such as temperature, concentrations, and duration have a significant effect on yield. The assumption that every reaction in the dataset is optimized for reaction parameters proved too loose, resulting in inaccurate predictive models for yield, and highlighting the importance of creating datasets with full (including potentially sub-optimal) reaction conditions.

More relevant to our work, \citet{perera2018platform} introduced a dataset of 5760 Suzuki-Miyaura cross-coupling reactions, \citet{ahneman2018predicting} introduced a dataset of 3956 Buchwald–Hartwig aminations, and \citet{prieto2022accelerating} investigated screening additives for Ni-catalysed reactions, all for the purposes of yield prediction. The datasets have been used in the benchmarking of Gaussian processes and Bayesian neural networks \citep{griffiths2023gauche}, deep learning models \citep{schwaller2021prediction}, language-model-based embeddings \citep{rankovic2025gollum}, data augmentation techniques \citep{schwaller2020data}, and Bayesian optimisation \citep{rankovic2024bayesian}. In each case, the datasets focus on discrete reaction variables, such as ligand, base, additives, or reactants at fixed temperatures and residence times. We are instead introducing a dataset rich in continuous reaction conditions (in our case temperature and residence time), as well as providing a pseudo-continuous representation of solvents themselves through the use of solvent mixtures.

Perhaps the closest example to our dataset is presented in \citet{nguyen2019high}, who used high-throughput experimentation to screen 12708 catalyst informatics on the oxidative coupling of methane. In this case, they do provide process conditions for temperature, reactant equivalents, and flow rates; however, they do so in a discretized manner, as opposed to our approach that produces denser continuous representations of variables. The dataset has been used in the context of benchmarking language models for yield prediction, where the variables are used to create prompts to generate LLM embeddings of reactions. Introduced by \citet{ramos2023bayesian}, the LLM embeddings are used for Bayesian optimization in reaction space. Recently, \citet{rankovic2025gollum} fine tune a subset of 1180 LLM embeddings to use as deep kernel GPs, achieving even more favorable performance.

More detailed and dense datasets including kinetic data usually have to be searched for in the research papers that originally published them, which are often accompanied by kinetic model fitting and benchmarking \citep{sagmeister2023accelerating, sagmeister2024simultaneous, silber2023accelerating}. However, the datasets are seldom ML-ready, and tend to focus on variables which have predictable outcomes. In this work, we collect solvent data, which has a very large impact on the system's dynamics and is often very challenging to model theoretically, making it a particularly interesting instance for machine learning applications.

\subsection{The dataset}

\begin{figure}[htbp]
    \centering
    \begin{subfigure}[b]{0.45\textwidth}
        \centering
        \includegraphics[width=\textwidth]{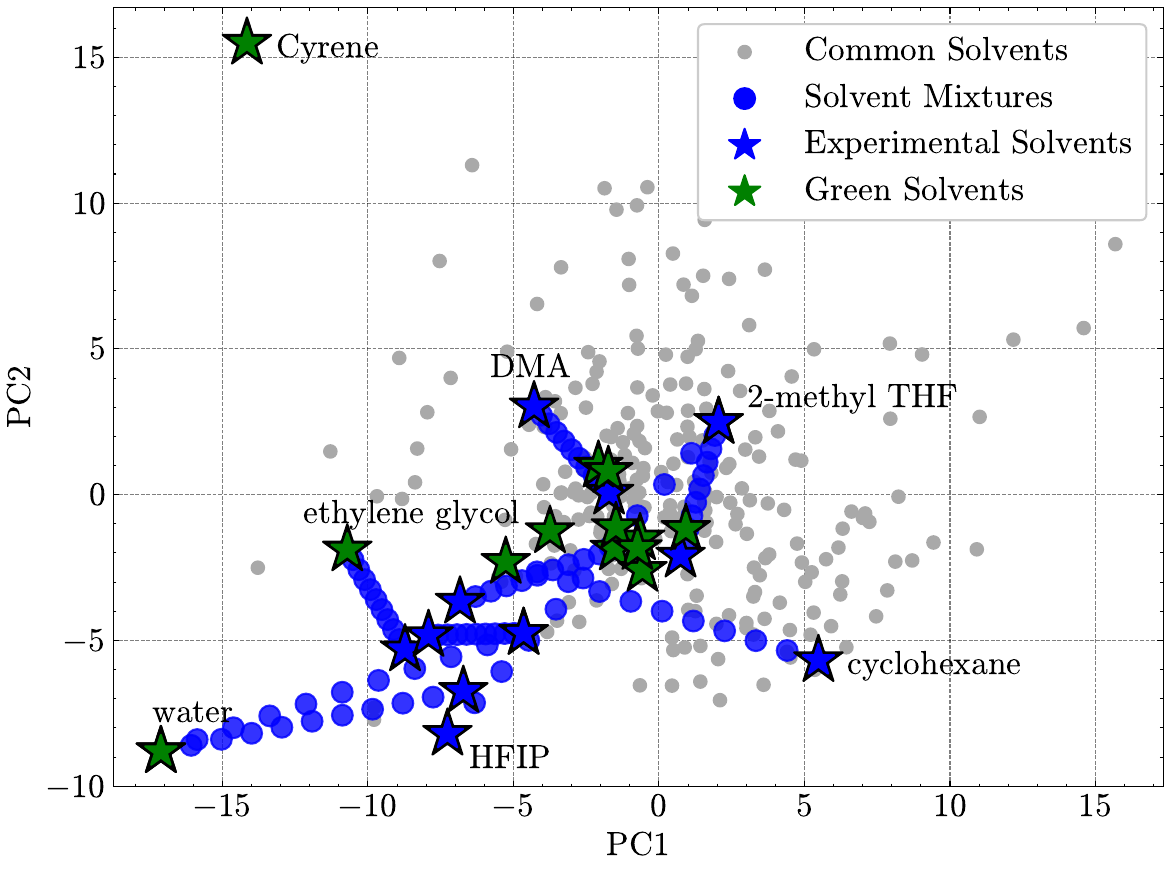}
        \caption{ACS PCA representation of the space of solvents. We highlight the solvents we collected data for with green or blue stars, and show the mixture solvents as dots.}
        \label{fig: solvent_space}
    \end{subfigure}
    \hfill
    \begin{subfigure}[b]{0.45\textwidth}
        \centering
        \includegraphics[width=\textwidth]{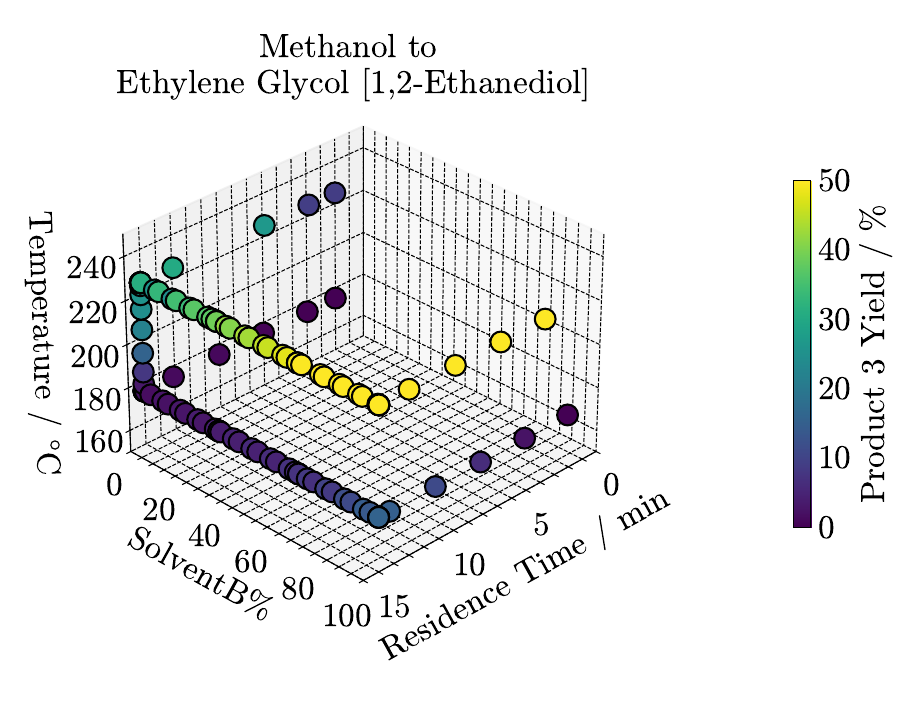}
        \caption{Three-dimensional scatter plot showing an experimental run between two solvents. We see examples of residence time ramps, temperature ramps, and solvent ramps.}
        \label{fig: surface_example}
    \end{subfigure}
    
    \vspace{0.5cm}
    
    \begin{subfigure}[b]{0.45\textwidth}
        \centering
        \includegraphics[width=\textwidth]{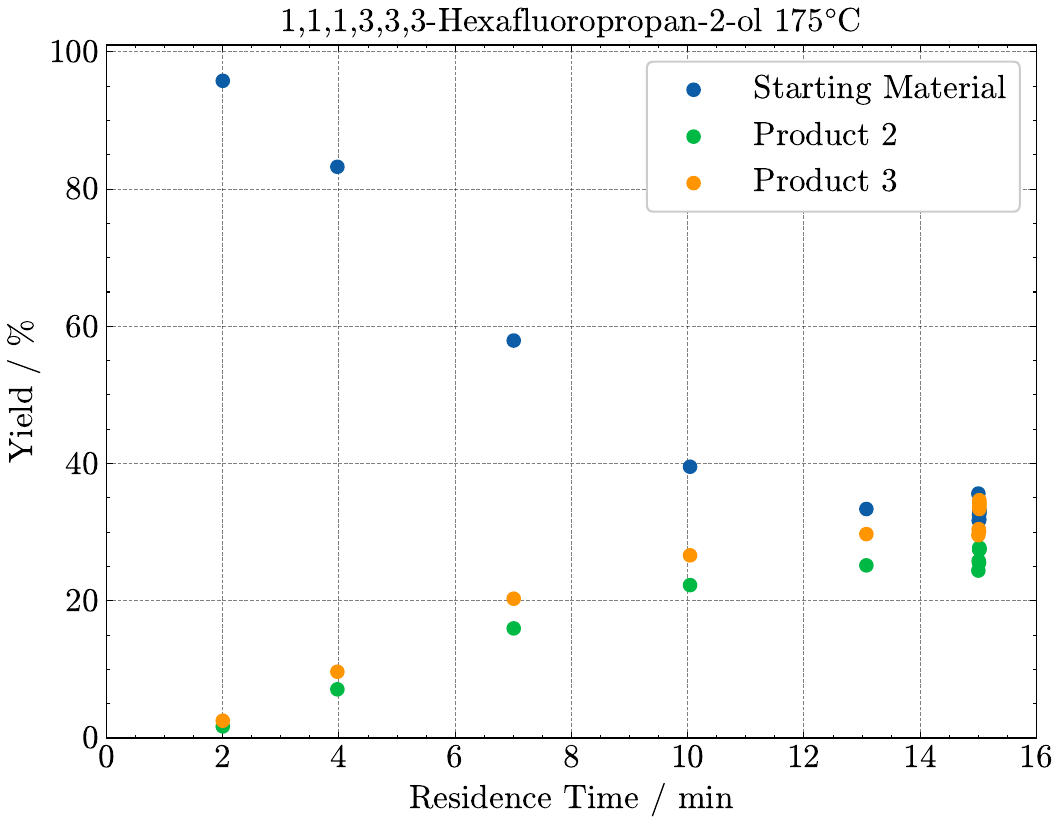}
        \caption{Example of a residence time ramp under the HFIP solvent. We see how longer reaction time increases product yield.}
        \label{fig: rt_ramp_example}
    \end{subfigure}
    \hfill
    \begin{subfigure}[b]{0.45\textwidth}
        \centering
        \includegraphics[width=\textwidth]{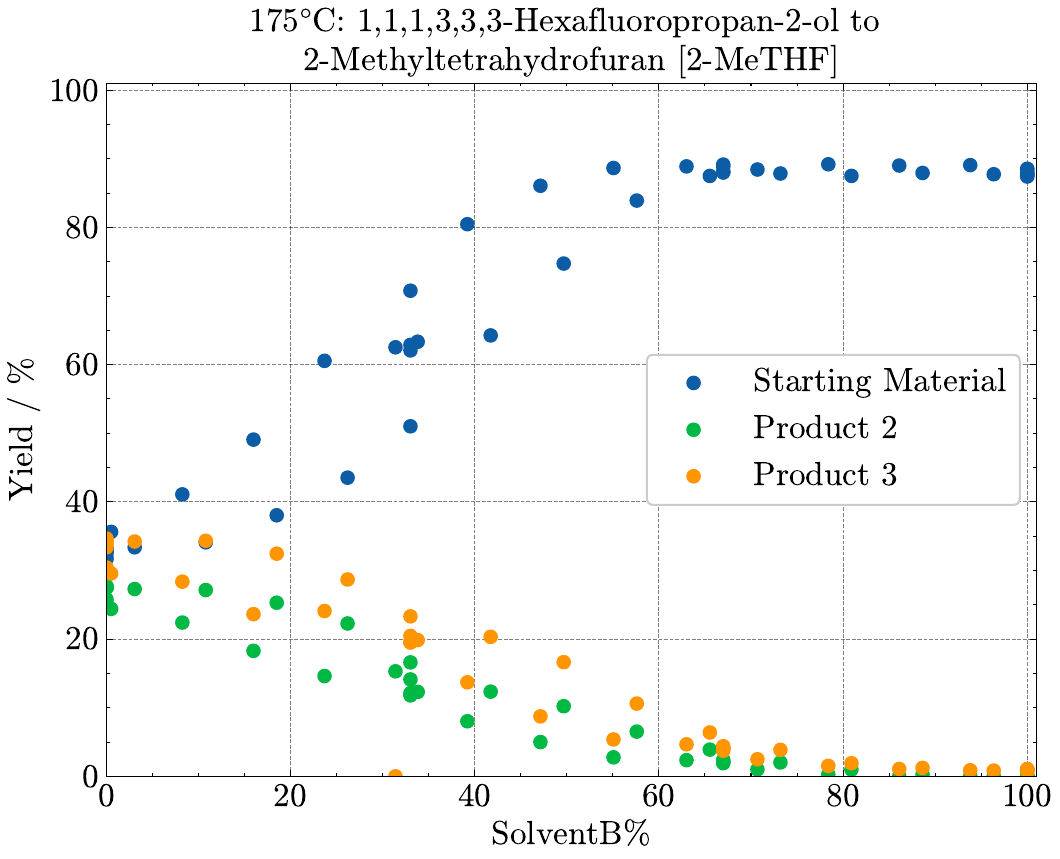}
        \caption{Example of a solvent ramp between HFIP and 2-MeTHF, exemplifying two of the challenges of the dataset: bias and heteroskedasticity.}
        \label{fig: solvent_ramp_exp}
    \end{subfigure}
    
    \caption{Visual summary of the data set. (a) Showcases the solvent space covered. (b) A full 8h experimental run between two solvents. (c) A residence time ramp, showing the starting material and product yields. (d) A solvent ramp, showing the yields under solvent mixture conditions.}
    \label{fig: visual_summary}
\end{figure}

The full dataset we collected for this project consists of 1227 data-points, with different reaction conditions, with the inputs being:
\begin{enumerate}
    \item[(1)] A selection of two different solvents in which the reaction was carried out, and the corresponding ratio of the solvents in the mixture.
    \item[(2)] The temperature at which the reaction was carried out.
    \item[(3)] The residence time of the reaction, i.e., how long the reactants were subject to the reaction conditions applied.
\end{enumerate}
The outputs consist of the yield of the starting material and the two observed products. We can further extract a dataset of single solvent data only. A visual summary of the data is provided in Figure \ref{fig: visual_summary}.

We further expand the dataset by including previous measurements on the same reaction class collected on a similar molecule, which was reported when developing the solvent ramping technology~\citep{WO2025073762}. The two datasets are detailed in Table \ref{tab: dataset_tab}, and can be downloaded from Kaggle \footnote{Dataset: \url{https://www.kaggle.com/datasets/aichemy/catechol-benchmark/}. \\ Code: \url{https://github.com/jpfolch/catechol_solvent_selection}.}. We further include sustainability details for all the solvents screened in Table \ref{tab: green_solvents_grouped} according to the GSK guide \citep{henderson2011expanding}.

\begin{table}[ht]
\centering
\renewcommand{\arraystretch}{1.5} 
\caption{Summary of the datasets: solvent types, data sizes, output measurements, and presence of time-series data. SM = Starting Material.}
\begin{tabular}{@{}llcccc@{}}
\toprule
\textbf{Dataset} & \textbf{Subset} & \textbf{Data Points} & \textbf{Solvents} & \textbf{Output Yields} & \textbf{Time-Series} \\
\midrule
\multirow{2}{*}{\makecell{Allyl Substituted \\ Catechol}} 
 & Solvent Mixtures & 1227 & 24 & SM + 2 Products & \checkmark \\
 & Single Solvents   & 656  & 24 & SM + 2 Products & \checkmark \\ \hline
\makecell{Allyl Phenyl \\ Ether} & Solvent Mixtures & 283 & 11 & SM + 1 Product & $\times$ \\
\bottomrule
\end{tabular}
\label{tab: dataset_tab}
\end{table}

\begin{table}[h!]
\centering
\caption{List of screened solvents and their classification. For more details see \citet{henderson2011expanding}.}
\label{tab: green_solvents_grouped}
\begin{tabular}{p{4cm}p{9cm}}
\toprule
\textbf{Classification} & \textbf{Solvents} \\
\midrule
\textcolor{green!60!black}{Green} & Ethylene Glycol; IPA; Water; Ethanol; Cyrene; Ethyl Acetate; DMC \\
\textcolor{orange!80!black}{Situation dependent} & Methanol; 2-MeTHF; Cyclohexane; Acetonitrile; Acetic Acid; 2-Butanone; t-Butanol \\
\textcolor{red!80!black}{Needs replacement} & Diethyl Ether; DMA; THF; MTBE \\
\textcolor{blue!80!black}{Not on GSK guide} & HFIP; 2,2,2-TFE; Decanol; Methyl Propionate; Ethyl $\ell$-Lactate \\
\bottomrule
\end{tabular}
\end{table}



\section{Dataset collection and techniques}
\label{sec: data_collection_and_techniques}

This section provides general descriptions of the data collection techniques, transient flow, analytical analysis, deconvolution, and how active learning was used for ramp selection.

\subsection{Transient flow and solvent ramping}

 Flow chemistry refers to a process in which the reaction is carried out in a continuous stream of reactant materials confined within tubing or narrow channels \citep{plutschack2017hitchhiker}. This technology offers an alternative to the traditional batch vessels typically used in chemical manufacturing and often presents benefits in areas such as safety, environmental impact, and scalability predictions \cite{hartman2011deciding}. Advantages include improved heat and mass transfer, more precise process control of the reaction conditions due to the smaller reaction volumes, and the ease of integration of online equipment and analytics \citep{boyall2024automated}.

\begin{figure}
    \centering
    \includegraphics[width=0.75\linewidth]{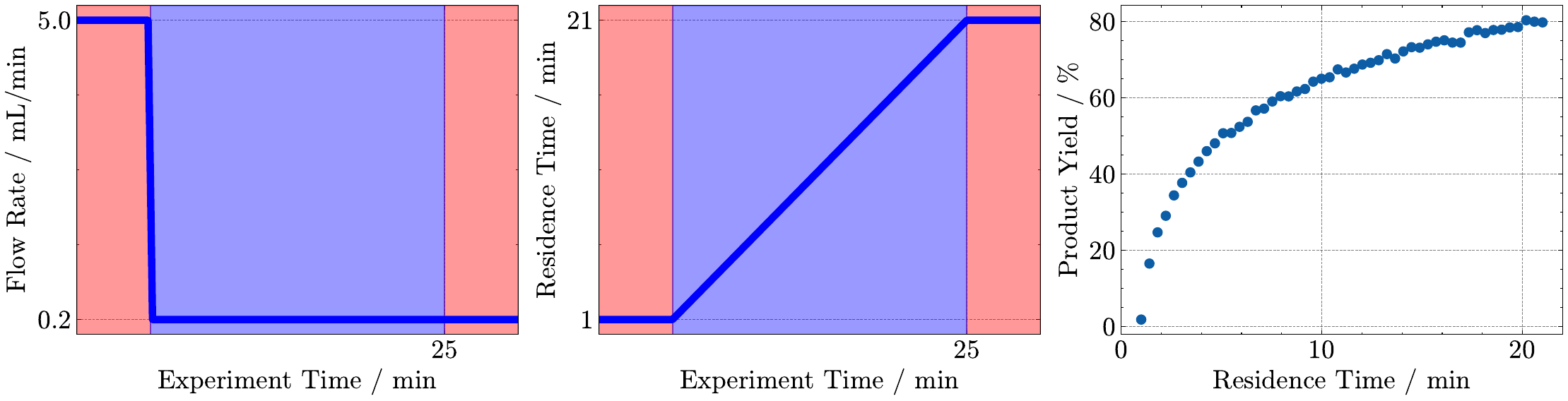}
    \caption{Example of a residence-time ramp in a transient flow reactor. (Left) We decrease the flow rate of the reactor to begin the experiments. (Middle) The residence time experienced by the flow at the point of measurement. (Right) Product yield mapped against residence time of measurements.}
    \label{fig: rt_ramp}
\end{figure}

Transient flow chemistry is an emerging technology used for collecting large quantities of reaction data in a continuous system. The method differs from traditional steady-state flow chemistry techniques, as the reaction conditions are varied continuously during experimentation to screen a range of reaction conditions \citep{williams2024dynamic}. Due to the efficient mixing inherent to flow systems, when the reaction conditions are adjusted in a controlled manner, each individual part of the flow is subject to different reaction conditions (i.e., plug flow), resulting in the efficient collection of a series of data \citep{schrecker2023discovery}. An example of a relationship that can be investigated using this technique is the effect of the reaction time (i.e., the residence time of the reaction) on the yield of the reaction. This can be done by changing the cumulative flow rate in the reactor. The flow rates of the system are initially set to correspond to a specific residence time, and a step change to lower flow rates is performed. This means that the plugs at the end of the reactor experiences a shorter residence time in the reactor, and each subsequent volume of flow, or `plug,' will have a longer residence time, producing a continuous data series of increasing residence times \citep{schrecker2024comparative}. A visualization of this process is given in Figure \ref{fig: rt_ramp}.

In a similar fashion, other variables can be investigated, such as temperature (varied by slowly ramping the reactor temperature)~\citep{moore2016kinetics}, and the equivalents of reagents in a reaction (by modifying the ratio of flow rates pumped from different reagent reservoirs)~\citep{mcmullen2023automated}. Reaction solvent is of particular interest in this dataset, following research interest in finding `green solvents' as alternatives to traditional solvents \citep{drelinkiewicz2024rapid, WO2025073762}. Solvent (mixtures) are treated as continuous variables, where the ratio of the two chosen solvents is varied using the ratio of the flow rates of the respective pumps to screen different solvent mixtures, and the changes in reaction yield are observed. Figure \ref{fig: solvent_space} shows all the solvent mixtures we were able to sample, as we effectively gather data between pure solvent pairs.

\subsection{Solvent selection}
In order to maximize the amount of information gathered from the data set, we used active learning. In particular, we trained a Gaussian process model on the Allyl Phenyl Ether dataset \citep{WO2025073762}, which was the first published dataset that investigated solvent ramping using transient flow. We then selected a range of suitable available solvents~\citep{diorazio2016toward} to create a set $\mathcal{S}$ and selected the solvents to query according to the entropy criterion:
\begin{equation}
    s_{A, n+1}, s_{B, n+1} = \argmax_{s_A, s_B \in \mathcal{S} \times \mathcal{S}} H( Y(s_A, s_B) | D_n ),
\end{equation}
where $H$ is the GP's entropy, $D_n = \{ X(s_{A, i}, s_{B, i}) \}_{i=1}^n$ the set of solvent ramps gathered up to time $n$, and $Y(s_A, s_B)$ the vector of data points gathered during solvent ramping from $s_A$ to $s_B$.

\subsection{Data acquisition and preprocessing}

Online analytical measurements were collected using high-performance liquid chromatography (HPLC) \citep{dixon2024operator}, sampled every two minutes. This allows quantitative yield measurements to be collected over the course of the reaction, which can then be related to the reaction conditions applied to each sample. These can be calculated for each variable since we know the reactor volume, the flow rates, the temperature, and the duration that each sample in the flow stream was subjected to particular reaction conditions. The residence time is calculated by considering the measurement time, $t_m$, the reactor volume, $V$, and the cumulative flow rate function, $F_c(t)$. We then solve the equation:
\begin{equation}
    V = \int_{t_i}^{t_m} F_c(t) dt 
\end{equation}
to find the initial time the plug entered the reactor, $t_i$. From this we can then estimate all the reaction conditions, residence time, $R_\tau$, solvent $B$ percentage, $S_{B\%}$, and the average temperature, $\hat{T}$:
\begin{equation}
    R_\tau = f_m - f_i ; \quad S_{B\%} = \frac{F_B(t_i)}{F_c(t_i)}; \quad \hat{T} = \frac{1}{t_m - t_i} \int_{t_i}^{t_m} T_r(t) dt
\end{equation}
where $F_B(t)$ is the flow rate of the solvent $B$ pump at time $t$, and $T_r(t)$ is the temperature of the reactor at time $t$. We generally seek to make small, incremental changes in temperature and flow rates to obtain accurate measurements.

\section{Machine Learning Benchmarks}

In this section, we train a variety of machine learning models to investigate the performance of standard state-of-the-art models for this prediction task. In particular, we examine a large range of solvent featurization methods, and algorithms that have shown strong performance in the past.

\subsection{Solvent featurization}
\label{scn:featurization}

Perhaps the most challenging, and most important, component of the benchmark problem we present is the solvent featurization process. This step asks how to represent each solvent (mixture), such that ML algorithms can extract suitable information for accurate predictions. 


As our goal is to predict yield surfaces on unseen solvents, a featurization that allows for transfer of information between solvents is required. \citet{diorazio2016toward} introduced a dataset of 272 solvents, with over 100 features for each, and further provide a 5-dimensional PCA representation of the solvent space. A second representation uses measurable properties of solvents~\citep{spange2021reappraisal}, allowing easy grouping of solvents by type, e.g., as esters, ethers, and alkanes.

Cheminformatic features of molecules \citep{landrum2006rdkit}, `fragments', are created using count vectors indicating the number of times a specific functional group appears in the molecule (following group contribution theory).  \citet{rogers2010extended} show that bit vectors indicating the presence of substructures in a molecule, coined `molecular fingerprints', can be used for molecular property prediction. We test the concatenation of both vectors, known as `fragprints' \citep{griffiths2022data}.

Finally, we investigate directly featurizing the reaction itself. This can be done, e.g., using the difference in sets containing molecular substructures in the starting materials and products \citep{schneider2015development, probst2022reaction}, known as differential reaction fingerprints (DRFP). Additionally, a reaction fingerprint can be learnt from larger open-source databases by using encoder-decoder neural networks \citep{schwaller2021prediction}, known as reaction fingerprints (RXNFP).

\paragraph{Featurizing solvent mixtures}
A further question of interest is that of how to represent mixtures of solvents. Given a pair of solvents and their respective featurizations $S_A$ and $S_B$, we will initially take the naive approach of using a weighted mean:
$
    S_{A\cup B}(b)=(1-b)S_A + bS_B,
$
where $b$ is the proportion of solvent B in the mixture. However, this linear transition can be an oversimplification of the underlying chemistry \cite{burrows2022solvation}, so we investigate learning a non-linear mapping in Section \ref{sec: gp_extensions}.



\subsection{Regression}
\label{scn:regression}

To evaluate the available machine learning tools for analyzing the dataset, we present a set of models for regression.  We regress on the solvent mixture and single solvent datasets described in Table \ref{tab: dataset_tab}. We perform leave-one-out cross validation for all the models. Further details on the methods used and experimental details can be found in the appendix.

When creating the test set, we take the mean of all repeated observations to avoid over-penalizing models that predict these reaction conditions poorly. We also omit the reactions containing acetic acid due to unintended a side-reaction%
\footnote{The presence of acetic acid and high temperatures resulted in an unintended side-reaction of the expected product - possibly an esterification - as soon as it formed, showing very little yield of desired product but with high conversion. As such, we removed the affected results from our benchmark numbers.}%
; creating models that are robust to unexpected side reactions poses an interesting future challenge.

\textbf{Gaussian processes} (GPs) are probabilistic, nonparametric models \citep{rasmussen2005gaussian}. They are characterized by a covariance function, or \textit{kernel}, that measures the similarity between a pair of inputs. These models provide uncertainty quantification, and perform well in low-data settings such as Bayesian optimization \citep{frazier2018tutorial}.

For our BaselineGP, we fit the data to the temperature and residence time inputs, ignoring the solvent, thus providing an improved proxy of one-hot encoding which has been shown to work well in low-data chemical regimes \cite{pomberger2022effect}. For the other GP methods, we transform the solvent into the featurized space, then use the Euclidean distance in that space as an input to an RBF kernel. In the appendix, we also evaluate a kernel over graphs, using pairwise shortest distances \citep{borgwardt2005shortest, xie2025bogrape, xie2025global}.

\textbf{Neural networks} are a now-ubiquitous tool for regression. We start with a simple multi-layer perceptron (MLP). We also fine-tune pretrained large language models (LLMs): ChemBERTa \citep{chithrananda2020chemberta}, a model pretrained on chemistry texts, and RXNFP \citep{schwaller2021prediction}, pretrained on chemical reaction classifications. This follows works that have shown LLM abilities to generalize across reactions using their string representations \citep{bran2024transformers}. In the appendix we further show results for neural process architectures which mimic GPs through meta-learning approaches \citep{garnelo2018conditional}.

\textbf{Latent ODE} methods use neural networks to model the latent state and dynamics of an ordinary differential equation (ODE) \citep{chen2018neural, qing2024system}, directly representing the underlying kinetics of the reaction. We also include an explicit state variant (LODE) \citep{rubanova2019latent}, and a neural ODE (NODE) with time-dependent dynamics.

The regression results in Table \ref{tab:fulldatafit} show the strength of the Spange featurization, which uses parameters known to be important to solvent effects. Whilst MLPs have a slight edge over GPs in their mean square error performance, the latter is also able to provide uncertainty quantification. Using LLM embeddings leads to poor performance, as reported previously, however, the same work reports strong performance when the embeddings are fine-tuned \citep{rankovic2025gollum} which is left to future work.

\begin{table}
   \centering
\caption{Regression performance on the full dataset. Mean squared error (MSE) and negative log predictive density (NLPD) are averaged across all leave-one-out data splits.}
\begin{adjustbox}{max width=\textwidth}
\begin{tabular}{llllrl}
\toprule
 &  & \multicolumn{2}{c}{Full data} & \multicolumn{2}{c}{Single solvent} \\
 &  & MSE ($\downarrow$) & NLPD ($\downarrow$) & MSE ($\downarrow$) & NLPD ($\downarrow$) \\
Model & Featurization  &  &  &  &  \\
\midrule
Baseline GP &  &  \textit{\textbf{0.011}} & -5.381 & 0.014 & -5.044 \\
\multirow[t]{4}{*}{GP} & acs  & 0.016 & -4.161 & 0.017 & -4.053 \\
 & drfps & 0.015 & -4.937 & 0.017 & -4.028 \\
 & fragprints & 0.013 & -5.010 & 0.017 & -4.481 \\
 & spange & \textbf{\textit{0.011}} & \textbf{-5.663} & \textit{\textbf{0.011}} & \textbf{-5.793} \\
 \cmidrule{1-6}
\multirow[t]{4}{*}{MLP} & acs & 0.014 & - & \textit{\textbf{0.011}} & - \\
 & drfps & 0.013 & - & 0.015 & - \\
 & fragprints & \textit{\textbf{0.011}} & - & \textbf{0.010} & - \\
 & spange & \textbf{0.010} & - & \textbf{0.010} & - \\
\multirow[t]{2}{*}{LLM} & rxnfp & 0.105 & - & 0.055 & - \\
 & chemberta & 0.153 & - & 0.074 & - \\
 \cmidrule{1-6}

NODE & spange & - & - & 0.055 & - \\
EODE & spange & - & - & 0.050 & -3.339 \\
LODE & spange & - & - & 0.049 & -3.235 \\
\bottomrule
\end{tabular}
\end{adjustbox}

    \label{tab:fulldatafit}
\end{table}

\begin{figure}
    \centering
    \includegraphics[width=0.8\linewidth]{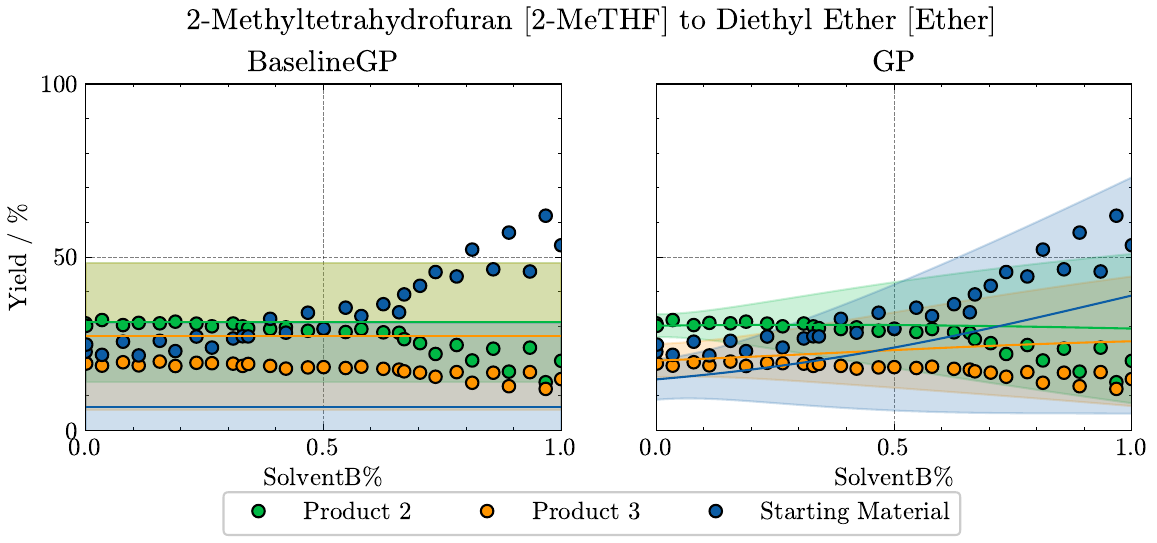}
    \caption{GP prediction on the yields of a solvent ramp, using Spange descriptors. We showcase a comparison between the baseline Gaussian process and a standard one. 2-MeTHF appears in another ramp, and so the model is confident about its predictions; as the proportion of Ether increases, so too does the model uncertainty.}
    \label{fig:gpprediction}
\end{figure}

\subsection{Gaussian process extensions} \label{sec: gp_extensions}

In Section \ref{scn:regression}, we report the performance of a selection of off-the-shelf machine learning algorithms for performing regression. However, many of these models fail to outperform the baseline model, which does not use any solvent information. We therefore propose some further GP models that can improve performance. Details of these models can be found in the appendix.

\textbf{Kernel design} must be performed carefully when creating GP models. A key issue with using the standard RBF kernel is that, for unseen solvents with featurizations very dissimilar to the solvents in the train set, the GP will revert to the uninformative prior. We therefore propose decomposing the kernel into solvent and non-solvent components in an additive manner, similarly to \citet{ru2020bayesian}.

\textbf{Multitask} kernels are able to learn correlations between outputs \citep{bonilla2007multi}. For example, the yields of the two products tend to be positively correlated with each other, and negatively correlated with the remaining starting material.

\textbf{Nonstationary} approaches allow modeling of search spaces that have changing lengthscale. For example, the rate of reaction is fastest in the first few minutes, and the solvent mixing may be nonlinear in the feature space as noted in Section \ref{scn:featurization}. We therefore learn a warping of these two inputs, inspired by \citet{snoek2014input} and \citet{balandat2020botorch}.

The results of these extensions are presented in Table \ref{tab:extensions}. These show promising directions to improve regression on single solvents, but struggle to beat the simpler GP when introducing solvent ramps.
We encourage the machine learning community to investigate further extensions, such as considering the heteroskedastic nature of the observation noise \citep{le2005heteroscedastic}, or non-stationary kernels \citep{van2017convolutional, boyne2025bark}.

\begin{table}
    \centering
    \caption{Regression performance of GP extensions on the catechol dataset.}
\adjustbox{max width=\textwidth}{
\begin{tabular}{llrrrr}
\toprule
 &  & \multicolumn{2}{c}{Full data} & \multicolumn{2}{c}{Single solvent} \\
 &  & MSE ($\downarrow$) & NLPD ($\downarrow$) & MSE ($\downarrow$) & NLPD ($\downarrow$) \\
Model & Extension &  &  &  &  \\
\midrule
BaselineGP &  & 0.011 & -5.381 & 0.014 & -5.044 \\
\multirow[t]{5}{*}{GP} &  & \textbf{0.011} & \textbf{-5.663} & 0.011 & -5.793 \\
 & Decomposed kernel & 0.012 & -5.455 & \textbf{0.009} & \textbf{-6.091} \\
 & Multitask GP & 0.018 & -2.885 & 0.011 & -2.494 \\
 & Input warping & 0.012 & -4.781 & 0.011 & -5.902 \\
\bottomrule
\end{tabular}
}
    \label{tab:extensions}
\end{table}

\subsection{Transfer learning} 

A key challenge in this dataset is the relatively low amount of data: where many modern ML approaches require large volumes of data, we only have 1227 observations of reaction conditions. To address this, we extend the training data by including results from the Ethyl dataset \citep{WO2025073762}, which has a further 283 experiments. For this regression problem, we only predict the total product yield, since the Ethyl dataset only has one observed product. 

We test the best performing regression models, the independent GPs and the MLP. For transfer of information across GPs, we use a multitask kernel corresponding to each reaction. For MLPs, we encode the reaction through a binary input. The baseline GP only uses the residence time and temperature, so cannot use the additional data. The results of this experiment are given in Table \ref{tab:transfer}, demonstrating the utility of learning across multiple datasets.

\begin{table}
    \centering
    \caption{Regression performance with transfer learning from the Ethyl dataset.}
\begin{tabular}{llrlll}
\toprule
 &  & \multicolumn{2}{c}{Catechol} & \multicolumn{2}{c}{Catechol + Ethyl} \\
 &  & MSE ($\downarrow$) & NLPD ($\downarrow$) & MSE ($\downarrow$) & NLPD ($\downarrow$) \\
Model & Featurization &  &  &  &  \\
\midrule
BaselineGP &  & \textbf{0.023} & -1.331 & 0.023 & -1.331 \\
GP & spange & 0.030 & \textbf{-1.487} & \textbf{0.020} & \textbf{-1.506} \\
MLP & spange & 0.027 & - & 0.031 & - \\
\bottomrule
\end{tabular}
    \label{tab:transfer}
\end{table}

\subsection{Active learning and Bayesian optimization}

\begin{figure}[htbp]
    \centering
    \begin{subfigure}[b]{0.45\textwidth}
        \centering
        \includegraphics[width=\textwidth]{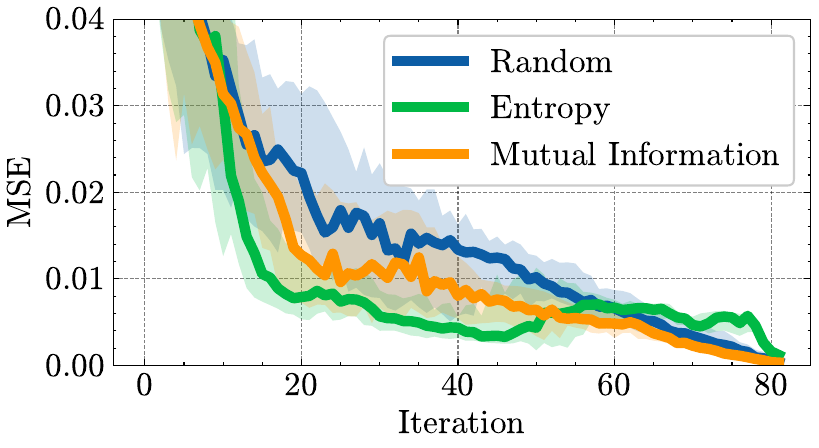}
        \caption{Active learning on transient flow ramps.}
        \label{fig: active_learning}
    \end{subfigure}
    \hfill
    \begin{subfigure}[b]{0.45\textwidth}
        \centering
        \includegraphics[width=\textwidth]{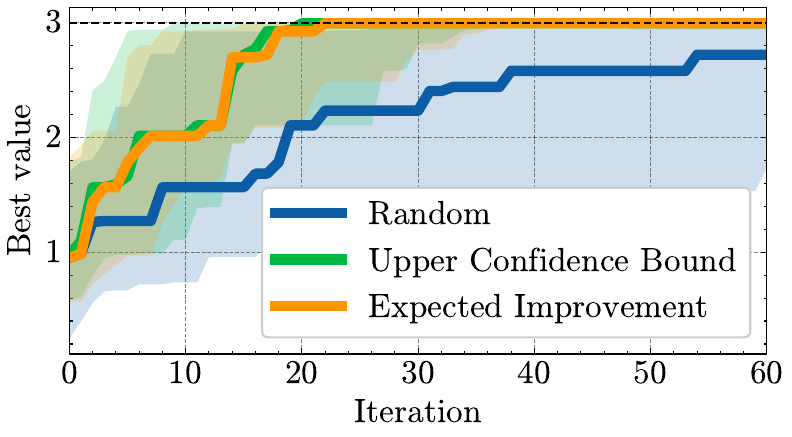}
        \caption{BO benchmarking across reaction space.}
        \label{fig: bayes_opt}
    \end{subfigure}
    
    \caption{Results of benchmarking for active learning and BO. We initialize the GP with 5 random samples, and show results over 30 initializations. We report the median, 10th and 90th quantiles.}
    \label{fig:my_2x2_gri}
\end{figure}

One key application for machine learning in chemistry is to optimally design experimental conditions, with recent interest in transient flow applications \citep{folch2022snake, mutny2023active, folch2024transition}. In this section, we showcase how the dataset can be used to benchmark algorithms for design of experiments. For simplicity, we focus on the independent GP model with descriptors from \citet{spange2021reappraisal}. We first explore \textit{active learning} ideas in transient flow: we split the dataset into ramps and use the entropy and mutual information criteria \citep{krause2007nonmyopic} to select transient ramps sequentially, with the goal of maximizing information across the dataset, which we measure by MSE. Figure \ref{fig: active_learning} shows that using the entropy criterion reduces MSE more initially, while using mutual information gives the best long-term performance.

We then benchmark on classical Bayesian optimization algorithms Expected Improvement \citep{mockus1998application} and Upper Confidence Bound \citep{srinivas2009gaussian}. We design an objective to maximize product yield and the selectivity of Product 2, while minimizing temperature and residence time, exemplifying conflicting objectives in the scale-up process. We allow the algorithms to query single points across the whole dataset, with the goal of identifying the optimal configuration in the fewest queries. Figure \ref{fig: bayes_opt} shows the results. In this case, we observe that the algorithms are very quickly able to identify the optimum, usually after 20 iterations, outperforming a random search by a significant margin.

Finally we consider a multi-objective optimization benchmark, by considering a three dimensional objective function of trying to optimize yield, selectivity, and a green-score created from Table \ref{tab: green_solvents_grouped}. We consider the solvent greenness by setting a value of 1.0 to every green solvent, -1.0 to every harmful solvent, and 0.0 to the remaining. For mixture solvent data-points we take a weighted average of their green scores. As benchmark metrics we consider three metrics of Pareto coverage: Euclidean generational distance (GD), inverted generational distance (IGD), and the maximum Pareto frontier error (MPFE) \citep{riquelme2015performance}. We present results in Table \ref{tab: mobo_results}, where we benchmark Thompson Sampling with random scalarazations \citep{paria2020flexible}.

\begin{table}[ht]
\centering
\caption{Results of multi-objective optimization benchmarking over iterations. We compare Euclidean generational distance (GD), inverted generational distance (IGD), and the maximum Pareto frontier error (MPFE). Multi-objective optimization results in much stronger metrics that randomly sampling.}
\label{tab: mobo_results}
\setlength{\tabcolsep}{6pt}
\renewcommand{\arraystretch}{1.1}
\begin{tabular}{c|cc|cc|cc}
\toprule
\multirow{2}{*}{\textbf{Iteration}} 
  & \multicolumn{2}{c|}{\textbf{GD} ($\downarrow$)} 
  & \multicolumn{2}{c|}{\textbf{IGD}($\downarrow$)} 
  & \multicolumn{2}{c}{\textbf{MPFE} ($\downarrow$)} \\
& \textbf{Random} & \textbf{MOBO} & \textbf{Random} & \textbf{MOBO} & \textbf{Random} & \textbf{MOBO} \\
\midrule
1   & 0.199 & \textbf{0.158} & 0.397 & \textbf{0.356} & 0.453 & \textbf{0.387} \\
25  & 0.107 & \textbf{0.043} & 0.177 & \textbf{0.146} & 0.421 & \textbf{0.225} \\
50  & 0.081 & \textbf{0.023} & 0.126 & \textbf{0.105} & 0.388 & \textbf{0.212} \\
75  & 0.069 & \textbf{0.017} & 0.106 & \textbf{0.084} & 0.385 & \textbf{0.200} \\
100 & 0.054 & \textbf{0.014} & 0.092 & \textbf{0.072} & 0.285 & \textbf{0.200} \\
\bottomrule
\end{tabular}
\end{table}


\begin{figure}
    \centering
    \includegraphics[width=0.4\linewidth]{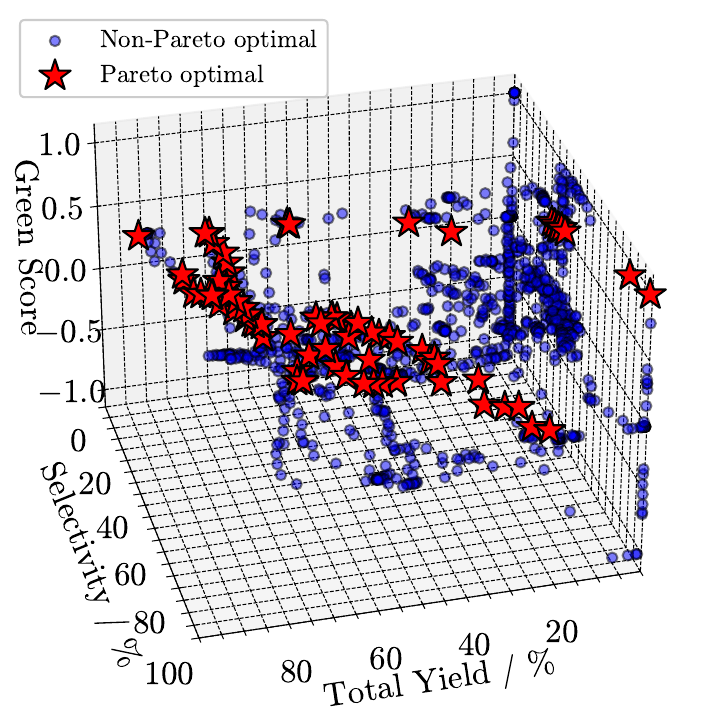}
    \caption{Visualization of empirical 3-dimensional Pareto front for the multi-objective Bayesian optimization benchmark.}
    \label{fig:placeholder}
\end{figure}

\section{Conclusions and future work}
\label{others}

This paper introduces the first ML-ready transient flow reaction dataset, showcasing the dynamic nature of chemical reactions not fully considered in past datasets. We particularly focus on solvent selection and challenge of learning solvent effects. We benchmark a variety of regression algorithms and solvent featurizations, many of which have shown strong performance in chemistry applications before. We show that many algorithms struggle in our dynamic setting due to a variety of factors: non-stationarity, hetero-skedasticity, and most importantly the lack of a good solvent featurization method. However, we show current techniques can still lead to important improvements, and can be effective in active learning settings.

We call on the machine learning community to develop improved methods for chemical dynamical systems. Such methods need to be ready to be infused with prior chemical knowledge, either through priors or data-driven learning. However, the most important step we must first address is the lack of data - truly effective predictive models will require large understanding of chemistry that cannot be obtained from single datasets. For example, some solvents may enable side reactions even when present only at small concentrations; as we observed in this dataset for the case of acetic acid. The best possible representation over mixed solvents should therefore reflect even trace amounts of these solvents, and then consider not only yield predictions, but the probability of the reaction actually happening. We hope this work enables an important next step for many ML researchers, to develop even more intelligent chemistry models in the near future.

Limitations of this paper include the focus on only two reactions, and while we touch on a large amount of machine learning models, we only go in depth with Gaussian processes due to their suitability to the small data regime. Future work would include improvements and further research into deep learning models, more flexible Bayesian models such as Bayesian neural networks, and investigating further ways of encoding chemical information into models.

\pagebreak

\section*{Acknowledgments}

The authors thank the AI for Chemistry: AIchemy hub for funding (EPSRC grants EP/Y028759/1 and EP/Y028755/1). We would further like to thank the Engineering and Physical Sciences Research Council (grants EP/W003317/1 to RM\&JCS, EP/X025292/1 to CT, RM, \& JQ, StatML CDT EP/Y034813/1 and IConIC EP/X025292/1 to TB \& BDL), a BASF/RAEng Research Chair in Data-Driven Optimisation to RM, a BASF/RAEng Senior Research Fellowship to CT, an Imperial College Department of Computing Scholarship to YX, and BASF SE, Ludwigshafen am Rhein funding to BDL, SZ, \& TB. RM holds concurrent appointments as a Professor at Imperial and as an Amazon Scholar. This paper describes work performed at Imperial prior to joining Amazon and is not associated with Amazon.

\bibliographystyle{unsrtnat}
\bibliography{references}


\appendix

\section{Details on the models and benchmarks}

\subsection{Benchmarking details}

\subsubsection{Regression}

For regression on the dataset, we perform leave-one-out cross validation. For the single solvents, we leave out one solvent at a time. For the full data, we leave out one solvent ramp at a time. We measure the performance of the model on each leave-one-out data split, then take the mean of their performance across the dataset. We exclude any experiments involving acetonitrile and acetic acid, due to the observed side-reactions. In addition, when considering the testing in single solvent data, we create a set of single data-points by averaging over repeated measurements, in order to remove mean error weighting from the longer residence times, in order to understand if the models catch the time-series nature of the data.

\subsubsection{Transfer learning}

As above, we perform leave-one-out cross validation on the solvent ramps in the catechol dataset. However, when we train each model, we append the training data from the ethyl dataset, alongside a binary feature indicating which dataset each observation is from. We also replace the three outputs of the catechol dataset (SM, Product 2, Product 3) with a single column, Product, which is the sum of the two products. This allows us to compare across the two datasets, since the ethyl dataset only has a Product column.

\subsubsection{Active learning and Bayesian optimization}

For Bayesian optimization we optimize the weighted objective function:
\begin{equation}
    f(S_A, S_b, b, \tau, T) = \lambda_1 (P_2 + P_3) + \lambda_2 \frac{P_2}{P_2 +P_3} - \lambda_3 \frac{T -175}{50} - \lambda_4 \tau
\end{equation}
where $S_A$ is solvent A, $S_B$ is solvent B, $b$ is the percentage composition of solvent B, $\tau$ is the residence time, $T$ the temperature, and $P_2$ / $P_3$ the yields of Products 2 and 3 respectively. We set the weight parameter values to:
\begin{equation*}
    \lambda_1 = 5; \quad \lambda_2 = 1; \quad \lambda_3 = 3; \quad \lambda_4 = \frac{1}{20}
\end{equation*}

For the Upper Confidence Bound acquisition function we use the standard exploration parameter $\beta = 1.96$. 

For locations with repeated measurements we simply consider average of all observations as the true product yields. All acquisition function optimizations are done through a simple exhaustive search of the space.

\subsection{Model details}

In this section, we provide the details necessary to reproduce the models used in the experiments. Any information that is not listed here can be found in our code, at \url{https://github.com/jpfolch/catechol_solvent_selection}.

\subsubsection{Gaussian processes}

We implement the GP models in this paper in BoTorch v0.13.0 \citep{balandat2020botorch}. We use the priors recommended by \citet{hvarfner2024vanilla}, to ensure good performance across featurizations of different dimensions. We use an RBF kernel, with the lengthscale prior
\begin{equation*}
    p(\ell) = \mathcal{LN}(\sqrt{2} + \log \sqrt{D} , \sqrt{3})
\end{equation*}

All GPs were trained using the MLII likelihood (maximum a posterior), with a training timeout of 30 seconds. For all of the GP extensions (in Table \ref{tab:extensions}), we use the Spange featurization.

\textbf{BaselineGP.} This model is a GP trained only using the residence time, and the temperature. This model does not factor in which solvent each experiment is from.

\textbf{DeepGP.} This model first trains a BaselineGP, then uses that as a mean function for another GP. In this way, far away from known solvents this model will fall back to the BaselineGP as a prior.

\textbf{Decomposed kernel.} We take inspiration from \citet{ru2020bayesian}, and separate our kernel into two parts. Specifically, we consider the input to the model to be the concatenation of the solvent featurization, $f$, and the non-featurized inputs, $x$, which include residence time and temperature. We then use the following kernel,
\begin{equation*}
    k_\text{decomp}([x, f], [x', f']) = k_x(x, x')\cdot k_f(f, f') + k_x(x, x') + k_f(f, f')
\end{equation*}
Similarly to the deep GP, this allows the features in $x$ to still contribute to the prediction, even when the unseen solvent is far from the known solvents.

\textbf{Multitask GP. } We use two different types of multitask GP in this paper. First, in Section 3.3, we use a multitask GP to represent each of the three measured yields. This kernel consists of a data kernel, and a task kernel,
\begin{equation*}
    k_\text{MT}([x, o], [x', o'])=k_x(x, x') \cdot k_o(o, o'),
\end{equation*}
where $k_o$ is an $O \times O$ matrix (for this dataset, $O=3$) that is used to learn the correlations between the outputs. Since all outputs are observed for each experiment, we can use a Kronecker structured kernel.

In Section 3.4, we use another multitask GP with 2 tasks, where each task corresponds to one of the two datasets. We use the same kernel as above, however only one task is observed at each reaction condition.

\textbf{Input warping. } In Section 3.3, we describe how the underlying chemistry is nonstationary. To attempt to address this, we take inspiration from \citet{snoek2014input} and \citet{balandat2020botorch}, learning a bijective map $\phi: [0, 1] \to [0, 1]$ that can capture the nonlinear effect of mixing solvents. This map has hyperparameters that can be learned,
\begin{equation*}
    S_{A \cup B}(b)=(1-\phi(b)) S_A + \phi(b)S_B, \qquad \phi(b)=1-(1-b^\alpha)^\beta,
\end{equation*}
where $\phi$ is the Kumaraswamy cumulative distribution function. We place a log normal prior on the parameters, $\alpha, \beta \sim \mathcal{LN}(0, \sqrt{0.3})$. This prior has median value of 1, which corresponds to a linear mapping.

\begin{figure}
    \centering
    \includegraphics[width=0.7\linewidth]{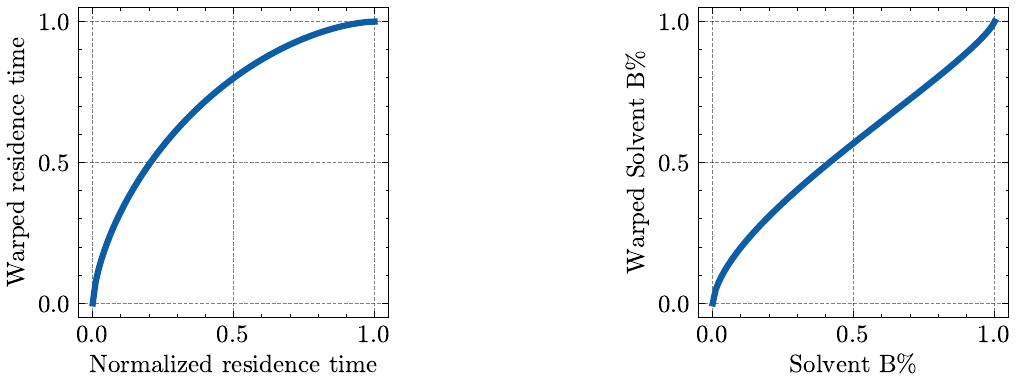}
    \caption{An example of a learned input warping, after training the GP on the full dataset.}
    \label{fig:inputwarping}
\end{figure}

We also use the input warping for the residence time. Since most of the reaction occurs in the first few minutes of the reaction, the lengthscale is far shorter compared to the later parts of the reaction. We find that this is indeed learned by the model, as shown in Figure \ref{fig:inputwarping}; the mapping effectively `spreads out' the observations early in the reaction, while compressing the later observations that tend to have a slower rate of change. Whilst the warping for the solvent composition learns a slight sigmoidal shape, we show experimentally in Section 3.3 that warping this feature does not improve regression performance.

\subsubsection{Neural networks}

Two types of neural network models were constructed for the regression tasks. The first was a standalone multilayer perceptron (MLP) model, and the second combined a large language model (LLM) backbone with an MLP head.

For the single-solvent task, the MLP model took as input the reaction time, temperature, and a feature vector representing the solvent. The network architecture consisted of two hidden layers with 128 and 64 neurons, respectively, each followed by ReLU activations and dropout (dropout rate of 0.5), and an output layer with 3 neurons.

For the mixed-solvent task, the MLP model used the same architecture, but the solvent input was computed as a sigmoid-weighted combination of the individual solvent feature vectors:

\[
S_{A\cup B} = \left(1 - \sigma_\theta(b)\right) S_A + \sigma_\theta(b) S_B,
\]

where \( S_A \) and \( S_B \) are the featurizations of solvents A and B, \( b \) is the percentage of solvent B in the mixture, and \( \sigma_\theta \) is a sigmoid function with trainable parameters \( \theta \).

The second model architecture used pretrained LLMs—\textbf{RXNFP} and \textbf{ChemBERTa}—to generate embeddings from reaction SMILES strings. For the single-solvent task, the SMILES representation of the reaction using the selected solvent was passed through the LLM to obtain the corresponding embedding. For the mixed-solvent task, the SMILES strings of the reactions carried out in solvents A and B, denoted \( \mathrm{RS}_A \) and \( \mathrm{RS}_B \), were each processed independently through the LLM to produce embeddings \( \mathbf{E}_A \) and \( \mathbf{E}_B \), respectively. These embeddings were then combined using a sigmoid-weighted sum:

\[
\mathbf{E}_{A \cup B} = \left(1 - \sigma_\theta(b)\right) \mathbf{E}_A + \sigma_\theta(b) \mathbf{E}_B,
\]

where \( b \) is the percentage of solvent B in the mixture and \( \sigma_\theta \) is a sigmoid function with trainable parameters \( \theta \). 

The resulting embedding was concatenated with the time and temperature, and passed through an MLP with the same architecture as the standalone MLP model. The LLM backbones were kept frozen during training, and only the MLP head was optimized.

The \texttt{ChemBERTa} model and tokenizer used were \texttt{seyonec/ChemBERTa-zinc-base-v1}, loaded via the Hugging Face \texttt{transformers} library. Similarly, the pretrained \texttt{RXNFP} model and tokenizer used are available from the \href{https://github.com/rxn4chemistry/rxnfp}{\texttt{rxnfp}} repository.

All models were trained using a learning rate of 0.001, a batch size of 32, for up to 400 epochs, or until reaching a maximum runtime of 720 minutes.

\subsubsection{ODE}

The ODE models were trained with a learning rate of 0.001, and 100 epochs. For the latent state and latent dynamics, we used a 32-dimensional space, and for all of the other representations we used a 64-dimensional space. Further information can be found in the provided code.

\subsection{Additional results}

We showcase additional results for Neural Processes \citep{garnelo2018conditional} and graph Gaussian processes \citep{borgwardt2005shortest, xie2025bogrape} in table \ref{tab:extra_results}.

\begin{table}
    \centering
    \caption{Regression performance on the single solvent dataset. Mean squared error (MSE) and negative log predictive density (NLPD) are averaged across all leave-one-out data splits. We include the shortest path kernel (sp) and the exponential shortest path kernel (esp).}
    \begin{adjustbox}{max width=\textwidth}
    \begin{tabular}{llrl}
    \toprule
    &  & \multicolumn{2}{c}{Single solvent} \\
    Model & Featurization & MSE ($\downarrow$) & NLPD ($\downarrow$) \\
    \midrule
    \multirow[t]{4}{*}{NP} & acs & 0.153 & -1.173 \\
     & drfps & 0.139 & -1.587 \\
     & fragprints & 0.135 & -1.495 \\
     & spange & 0.089 & -1.472 \\
     \cmidrule{1-4}
    \multirow[t]{2}{*}{GraphGP}  & sp & 0.046 & 2.464 \\
     & esp & 1.068 & 2.453 \\
    \bottomrule
    \end{tabular}
    \end{adjustbox}
    \label{tab:extra_results}
\end{table}

\section{Details on data collection}

\subsection{Reactor details}
Here we include the reactor and detail procedures.

\begin{figure}
    \centering
    \includegraphics[width=0.8\linewidth]{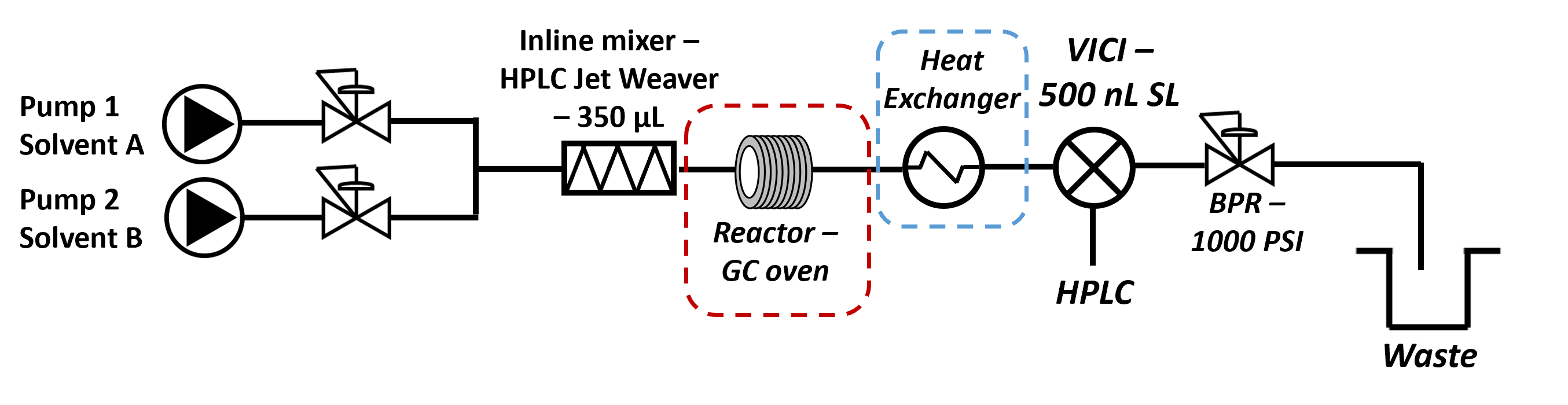}
    \caption{Piping \& instrumentation diagram of the automated continuous flow coiled reactor used to collect the transient flow data reported in this paper.}
    \label{fig:reactor}
\end{figure}

The automated reactor setup used to collect the data is shown in Figure \ref{fig:reactor}. Knauer Azure 4.1S pumps fitted with stainless steel 10 mL pump heads were used as pumps 1 and 2. All tubing used for the entire reactor was made of 316 stainless steel (1.5875 mm OD, 1 mm ID). An Agilent inline jet weaver HPLC mixer (350 $\mu$L volume) was used as an inline mixer to ensure the reactant solution was homogeneous before entering the reactor. An Agilent 6890 GC oven was used to heat the stainless steel coiled reactor (1.5875 mm OD, 1 mm ID, 7.95 mL volume) during the reaction to the desired temperature. A customized cooling system made from an aluminum block and a Peltier assembly was then placed inline to rapidly cool the flow of solution and quench the reaction. A Vici four port-2 position sampling valve followed the Peltier to sample small aliquots (500 nL) into the HPLC for online analysis measurements of the reaction. An IDEX 1000 PSI BPR was then placed before the waste tubing of the reactor to depressurize the reaction solution back to atmospheric pressure. The pumps, oven and Vici valve were automated by code developed in house in Python.

\subsubsection{Methods}
A typical reaction run was performed as following:
\begin{itemize}
    \item[1.] The reactant solutions were made up by adding allyl phenyl ether (50 $\mu$L) and the internal standard - ethyl benzene (50 $\mu$L) in to both solvent A and solvent B (250 mL) in separate volumetric flasks. \\
    \item[2.] The reactant pumps were primed with their respective solvents and pumped through the system at 1 mL min$^{-1}$ for 15 minutes.\\
    \item[3.] The pumps were then primed with the reactant solutions and pumped through the system at 1 mL min$^{-1}$ for 5 minutes. \\
    \item[4.] The HPLC was started and a sequence was created to record external sampling via the Vici Valve.\\
    \item[5.] The python code that runs the experiments was then initialized and the experiment was started. \\
    \item[6.] Once the reaction run was completed, the reactor is flushed with their respective solvents for 10 minutes at 1 mL min$^{-1}$, followed by a flush of the system with a miscible solvent (usually IPA) and cleaned for the next reaction. The data was stored in a SQL database and is then deconvolved offline.
\end{itemize}

All the data-points recorded were reported in the dataset, and the only outliers that were removed were those slugs that experienced a step-up in flow-rate while in the reactor, as this has been shown to add bias to the data \citep{schrecker2024comparative}.

\subsection{Fine-tuning calibration via optimization}

The HPLC data we obtained is uncalibrated, which means we cannot calculate yields directly from the peak areas collected from online HPLC measurements. However, the yields of each product follows the linear relationship with peak areas:
\begin{equation}
    y_{product} = \epsilon_{product} \times \frac{c_{IS}}{c_0} \times \text{peak\_ratio}
\end{equation}
where $c_{IS}$ is the internal standard concentration in mol L$^{-1}$, $c_0$ is the initial concentration of starting material in mol L$^{-1}$, and $\epsilon$ is the \textit{calibration constant}. The peak\_ratio refers to value given by dividing the area of the peak of interest (starting material, product 2 or product 3) by the peak area of the internal standard. This constant is calculated by performing calibrations of the HPLC detector with injections of pure compounds at different concentrations, while keeping the internal standard concentration constant, and therefore observing the linear relationship and obtaining the response factor of the compounds. Obtaining a pure sample of Product 2 and Product 3 however, turned out to be particularly difficult due to the compounds being isomers, making the separation of the pure products tough. Therefore, we instead focused on using the estimates we had and then fine-tuning them via an optimization procedure.

Our initial HPLC tests gave us the following estimates:
\begin{align*}
    \hat{\epsilon}_{SM} = \frac{1}{1.5} ; \quad \hat{\epsilon}_{P2} = \frac{1}{3}; \quad \hat{\epsilon}_{P3} = \frac{1}{3}
\end{align*}
From here, we decided to fine-tune the estimates in order for the calculated yields to ensure the reaction yields were mass balanced. We identified specific measurements where we expected full conversion (i.e. the sum of yields should be 100), and we further allowed for experimental concentrations to vary according to the error in the laboratory analytical pipettes used for making the reactant solutions. This results in the following optimization problem, where we penalized deviation from our initial calibration measurements, and deviation from full conversion at specified measurements $\mathcal{K}$:

\begin{align*}
\min_{\{c_{i},\, \epsilon_j\}} \quad &
\alpha \sum_{i} (c_{i} - 2.25)^2
+ \beta \sum_{j} (\epsilon_j - \hat{\epsilon}_j)^2
+ \gamma \sum_{k \in \mathcal{K}} \left( \sum_{j} y_{kj} - 100 \right)^2 \\
\text{where} \quad &
y_{ij} = \text{const} \cdot \text{peak\_ratio}_{ij} \cdot \epsilon_j \cdot c_{i}, \quad \forall i = 1, ..., 1227; j \in \{ SM, P2, P3 \} \\
& c_{i} = c_{i'} \quad \quad \text{ if } i, i' \text{ are in the same experimental run } \\
\end{align*}
with constraints to restrict total yield under 100\% and possible errors in concentrations:
\begin{align*}
& \sum_j y_{ij} \leq 100, \quad \forall i \\
& c_{i} \in [1.25, 2.5], \quad \forall i \\
& 0.2 \leq \epsilon_j \leq 0.5, \quad \forall j
\end{align*}

\noindent
where:
\begin{itemize}
  \item $c_{i}$ are the corrected concentration ratios,
  \item $\epsilon_j$ are the calibration scaling factors for each compound,
  \item $\text{peak\_ratio}_{ij}$ are the observed HPLC peak area ratios,
  \item $\mathcal{K}$ is the set of indices where full conversion is expected,
  \item $\alpha$, $\beta$, and $\gamma$ are weighting parameters.
\end{itemize}
we optimized with $\alpha = \beta = \gamma = 1$, optimized using \texttt{scipy}'s \texttt{minimize} function with the Sequential Least Squares Programming (SLSQP) algorithm. To select the initial values, we used a 100,000 initial grid search. This resulted in the following parameter estimates:
\begin{align*}
    \epsilon_{SM} = 0.525 ; \quad \epsilon_{P2} = 0.222; \quad \epsilon_{P3} = 0.361
\end{align*}

\subsection{Spange descriptor interpolation}
The descriptors from \citet{spange2021reappraisal} were obtained from the supplementary material on the paper. However, there are a few values missing from some rows, including for the solvents we gathered data for. In order to estimate the missing values, we trained a multi-task Gaussian process model on the whole table, under a Taniamoto kernel, which we then used to predict the missing values that are used for all the main methods in the paper.
\



\end{document}